# A Vision-Language Model for Focal Liver Lesion Classification

Jian Song[1,2], Yuchang Hu[1], Hui Wang[3] and *Yen-Wei Chen[2]

[1] School of Mathematical Sciences, Huaqiao University, Fujian, China
[2] College of Information Science and Engineering, Ritsumeikan University Osaka, Japan
[3] School of Information Science and Engineering, Shandong University, Qingdao, China

*Corresponding Author: chen@is.ritsumei.ac.jp

**Abstract.** Accurate classification of focal liver lesions is crucial for diagnosis and treatment in hepatology. However, traditional supervised deep learning models depend on large-scale annotated datasets, which are often limited in medical imaging. Recently, Vision-Language models (VLMs) such as Contrastive Language-Image Pre-training model (CLIP) has been applied to image classifications. Compared to the conventional convolutional neural network (CNN), which classifiers image based on visual information only, VLM leverages multimodal learning with text and images, allowing it to learn effectively even with a limited amount of labeled data. Inspired by CLIP, we propose a Liver-VLM, a model specifically designed for focal liver lesions (FLLs) classification. First, Liver-VLM incorporates class information into the text encoder without introducing additional inference overhead. Second, by calculating the pairwise cosine similarities between image and text embeddings and optimizing the model with a cross-entropy loss, Liver-VLM effectively aligns image features with class-level text features. Experimental results on MPCT-FLLs dataset demonstrate that the Liver-VLM model outperforms both the standard CLIP and MedCLIP models in terms of accuracy and Area Under the Curve (AUC). Further analysis shows that using a lightweight ResNet18 backbone enhances classification performance, particularly under data-constrained conditions.

## 1 Introduction

Focal liver lesions (FLLs) are a common finding in multi-phase computed tomography (CT) images, ranging from benign conditions such as hepatic cysts to malignant tumors like hepatocellular carcinoma (HCC). A multi-phase CT imaging typically consists of a Non-Contrast (NC) phase image, an Arterial (ART) phase image and a Portal Venous (PV) phase image. An example of different FLLs (i.e., Cyst, Focal Nodular Hyperplasia (FNH), Hepatocellular carcinoma (HCC), and Hemangioma (HEM)) in these multi-phase CT scans (NC, ART and PV) is shown in the Fig. 1. As shown in Fig.1, FLLs have different visual characteristics at different time points (phases) before/after intravenous contrast injection.

Accurate classification is critical for effective clinical decision-making. However, the examination of multi-phase CT imaging for diagnosing FLLs is a time-consuming and subjective task. Several deep learning-based networks [1-2] have been developed for the classification of FLLs. Although deep learning methods have achieved state-of-the-art performance, the limited availability of annotated medical images remains a barrier to further improving classification accuracy. In our previous research, we demonstrated that using pretrained deep learning models with ImageNet [3] or self-supervised learning [4-6] can substantially enhance the classification accuracy of FLLs. With the advancement of Large Language models and multimodal technologies, AI-assisted diagnosis has shown great potential in medical image analysis. However, its reliance on large annotated datasets remains a significant challenge. Drawing inspiration from the capability of the Contrastive Language-Image Pre-training (CLIP) model to learn generalized visual representations from textual annotations, we propose Liver-VLM, a text-guided framework for FLLs prediction.

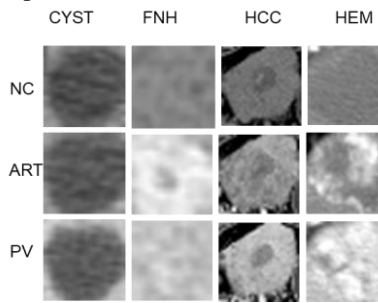

Fig. 1 Evolutionary Patterns of Four FLL Types Across Three Phases.

This paper is structured as follows: Section 2 provides a brief overview of related work. In section 3, we present a detailed description of the proposed approach. The experimental setup and results are presented in Section 4 and the concludes in the final section.

## 2    Related Work

### 2.1    Multimodal Pre-training

Multimodal pre-training enables models to learn transferable representations from datasets comprising multiple modalities, such as images, text, and audio. Unlike unimodal learning, it captures complex cross-modal relationships, enhancing downstream task performance. Seminal works have laid the foundation for multimodal pre-training. VirTex [7] learns image-text relationships for autoregressive caption generation, while ICMLM [8] employs masked language modeling to improve textual understanding and generation. Contrastive learning also plays a key role, as seen in ConVIRT [9] and CLIP [10], which align image-text pairs via contrastive loss to maximize positive pair similarity and minimize negative pair similarity. These approaches facilitate cross-modal knowledge transfer without extensive labeled data, benefiting applications such as medical imaging.

### 2.2 VLM Model

Vision-language models (VLMs) integrate vision and text for tasks like image captioning (BLIP [11]), visual question answering (LLaVA [12]), and image classification (CLIP [10], MedCLIP [13]). Unlike traditional models that rely solely on visual features, VLMs use text to enhance representation, especially in data-scarce settings. CLIP [10] learns a joint image-text embedding via contrastive learning, enabling zero/few-shot learning. Inspired by CLIP, MedCLIP aligns radiology reports with medical images for improved retrieval and classification. However, CLIP's reliance on web data limits its medical use, while MedCLIP, trained on X-rays, faces domain gaps in CT imaging. To bridge this gap, we propose Liver-VLM for FLL classification in multi-phase CT, employing cross-entropy loss to align image and class-specific text features, enhancing discriminative power for accurate classification.

## 3 Method

### 3.1 Overview of the Proposed Method

Fig. 2 provides an overview of the proposed Liver-VLM framework, which consists an image encoder (ResNet50 or ResNet18 [14]) and a text encoder (BERT [15]). The image encoder is designed to process multi-phase CT scans and extract visual embeddings, which are subsequently passed through a fully connected (FC) layer to match the dimensionality of the text embeddings. The similarity between image and text embeddings is measured using cosine similarity, enabling alignment within a shared embedding space. The image encoder is trainable during the training phase but remains frozen during inference to reduce computational overhead and mitigate overfitting. The text encoder maps expert-defined descriptions (e.g., class labels) into the same embedding space. By leveraging textual prompts in a supervised learning setting, the model strengthens the alignment between visual and textual representations, thereby enhancing the generalization capability. Due to the limited annotations typically available in medical imaging datasets, the pre-trained text encoder remains frozen during training, with only the fully connected ($FC_T$) layer appended after BERT being updated. By calculating the pairwise cosine similarities between image and text embeddings and optimizing the model using cross-entropy loss, the training objective seeks to maximize the similarity between each image and its corresponding text description, thereby ensuring effective alignment of visual and textual features. We propose two variants of the Liver-VLM model: one with the image encoder trained from scratch and the other with the image encoder fine-tuned.

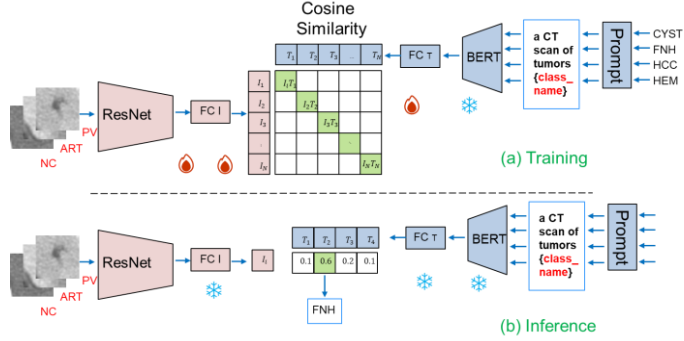

Fig. 2 The Workflow of Liver-VLM

### 3.2  Train from scratch

In the training-from-scratch model, the image encoder (ResNet50 or ResNet18) is randomly initialized and learns features solely from the multi-phase CT dataset without pre-trained parameters. The text encoder is a pre-trained BERT that remains frozen, with only the $FC_T$ layer being trainable. Using a supervised learning framework, the image encoder extracts key CT features and aligns them with frozen BERT-generated text embeddings via similarity-based optimization. A supervised loss function improves feature alignment, enhancing representation learning. The workflow of train from scratch is shown in Fig. 3.

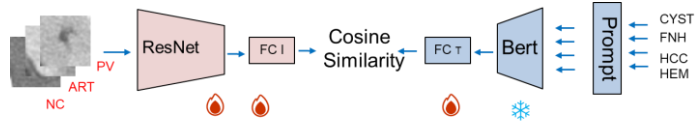

Fig. 3 The Workflow of Train from Scratch Liver-VLM

### 3.3  Fine-tuning

In the fine-tuned model, we first use an image encoder (ResNet50 or ResNet18) pre-trained on ImageNet as the image encoder. This allows the model to learn general visual features from a large-scale natural image dataset, providing a strong initialization for subsequent fine-tuning. Then, the pretrained image encoder weights are updated using the multi-phase CT dataset, enabling the model to adapt to the unique characteristics of medical imaging data. Since ImageNet primarily contains natural images, its pretrained features differ from those required for medical imaging, making the fine-tuning process essential. Similar to the training-from-scratch model, the text encoder, pre-trained BERT, remains frozen, with only the $FC_T$ layer after BERT being trainable. The model employs a supervised learning approach to optimize the alignment between learned image embeddings and frozen text embeddings, thereby enhancing classification performance. The fine-tuning workflow is shown in Fig. 4.

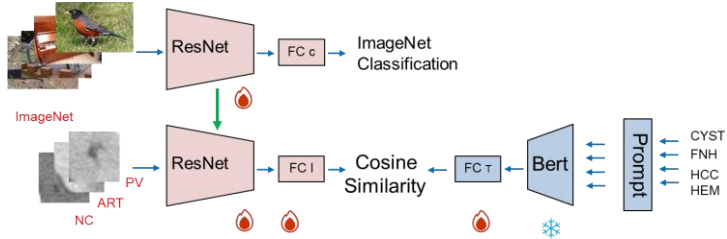

Fig. 4 The Workflow of Fine-tuning Liver-VLM

### 3.4 Prompt Engineering and Loss Function

To mitigate errors arising from ambiguous label texts lacking contextual information, we adopted prompt engineering techniques similar to those used in the CLIP model, constructing complete sentence texts using prompt templates. Given that liver lesion datasets predominantly utilize CT or MRI images, our CT image prompt template is "*a CT scan of tumors {label}*". Furthermore, to enhance textual description diversity in cases with a limited number of liver lesion categories, we expanded abbreviated lesion labels to their full names. For example, 'FNH' is represented as 'Focal Nodular Hyperplasia', 'HCC' as 'Hepatocellular Carcinoma', and 'HEM' as 'Hemangioma'.

Due to the small size of medical image datasets, the limited number of lesion classes, and the subtle semantic differences between categories, we choose to use the standard cross-entropy (CE) loss rather than the contrastive loss used in CLIP. The CE loss function is formulated as Equation (1):

$$\mathcal{L} = -\frac{1}{N}\sum_{i=1}^{N}\sum_{c=1}^{C} y_{i,c} \log(p_{i,c}) \qquad (1)$$

Where $N$ is the number of samples, and $C$ is the number of classes of FLLs. $y_{i,c}$ is the true label indicating whether the $i$th sample belongs to class $c$; if it does, then $y_{i,c} = 1$, otherwise $y_{i,c} = 0$. $p_{i,c}$ is the predicted similarity between the $i$th sample and class $c$.

## 4 Experimental Results

### 4.1 Dataset and Evaluation Metrics

We evaluate Liver-VLM on our in-house Multi-Phase CT dataset of Focal Liver Lesions (MPCT-FLLs) [16]. The dataset includes four types of lesions: Cyst, FNH, HCC, and HEM, collected from Sir Run Run Shaw Hospital, Zhejiang University, between 2015 and 2017. In total, the dataset comprises 85 CT volumes, corresponding to 489 selected slice images. For each volume, a set of slices centered on the lesion is chosen. The slice thickness is 5 or 7 mm, with an in-plane resolution of 0.57–0.59 mm. Each 2D slice image has a resolution of 512×512 pixels and consists of three CT phases: NC, ART, and PV. The regions of interest (ROIs) for each lesion were annotated by experienced radiologists. For experiments, the extracted 2D ROI slice images were resized to 128×128 pixels. The three-phase images were treated as a three-channel color image, resulting in an input image size of 3×128×128. Table 1 shows the distribution of data among three groups used for 3-fold cross-validation during model evaluation. In each fold, one group was designated as the test set, while the other two were used for training. The final performance metrics, including accuracy and AUC-ROC, were calculated by averaging the outcomes from all three folds.

**Table 1.** Dataset Distribution

|  |  | CYST | FNH | HCC | HEM | Total |
|---|---|---|---|---|---|---|
| Train: | case | 10 | 6 | 5 | 6 | 27 |
|  | slice | 59 | 33 | 29 | 30 | 151 |
| Test: | case | 10 | 5 | 7 | 7 | 29 |
|  | slice | 41 | 23 | 68 | 28 | 160 |
| Val: | case | 10 | 5 | 7 | 7 | 29 |
|  | slice | 49 | 20 | 65 | 44 | 178 |
| Total: | case | 30 | 16 | 19 | 20 | 85 |
|  | slice | 149 | 76 | 162 | 102 | 489 |

### 4.2 Implementations

For the pre-training, we train our network for 200 epochs with a batch size of 32 and a learning rate of 0.01. The AdamW [17] optimizer is used. The training setup is shown in Table 2.

**Table 2.** Computation Environment

| GPU | NVIDIA GeForce RTX 2060 |
|---|---|
| CPU | Intel(R) Core (TM) i7-10875H @2.30GHz |
| OS | Windows 10.0.26100.3194 |
| Deep Learning Framework | Pytorch |

### 4.3 Results

We conducted experiments on the MPCT-FLLs dataset for FLL classification, and the results are summarized in Table 3. A 3-fold cross-validation was performed using a CLIP model [10] with ResNet50 as the image encoder, which served as the baseline for comparison. This model achieved an average inference accuracy of 32.80±5.89% and an AUC of 0.40±0.03; however, the model classified all images as either CYST or HCC, with FNH and HEM achieving an accuracy of 0%. The MedCLIP model [13] was also evaluated on the same dataset. While its average accuracy (30.73±5.25%) and AUC (0.51±0.02), the model classified all images as either CYST or HEM, resulting in an accuracy of 0% for FNH and HCC.

Training ResNet50 from scratch (Model 1) significantly improved the average accuracy to 71.67±3.54% and the AUC to 0.88±0.02. Initializing ResNet50 with ImageNet pretraining before fine-tuning (Model 2) resulted in a 2.41% increase in average accuracy, although the AUC experienced a slight decrease of 0.01. Notably, the classification accuracy for the HEM category improved by 19.18%, reaching 63.15 ± 25.12 in Model 2 compared to 43.97 ± 13.41 in Model 1. However, the accuracies for CYST, FNH, and HCC in Model 2 were slightly lower than those in Model 1 by 1.38%, 0.35%, and 3.37%, respectively. Despite these variations, both Model 1 and Model 2 were able to correctly classify all four lesion types. The results demonstrate that our proposed model effectively captures generalized features, ultimately enhancing its predictive performance on small-sample datasets with limited annotations.

**Table 3.** Results on the MPCT-FLLs Dataset Using ResNet-50 as the Image Encoder

|  | CYST | FNH | HCC | HEM | Avg. Acc | AUC |
| --- | --- | --- | --- | --- | --- | --- |
| CLIP [10] | 26.15±11.70 | 0.00 | 75.37±2.32 | 0.00 | 32.80±5.89 | 0.40±0.03 |
| MedCLIP [13] | 69.42±2.36 | 0.00 | 0.00 | 45.16±9.30 | 30.73±5.25 | 0.51±0.02 |
| Model1 | **97.51±1.78** | **71.64±4.87** | **67.86±14.16** | 43.97±13.41 | 71.67±3.54 | **0.88±0.02** |
| Model2 | 96.13±1.09 | 71.29±14.47 | 64.49±10.56 | **63.15±25.12** | **74.08±4.33** | 0.87±0.02 |

Considering that the MPCT-FLLs dataset contains only 489 slices of multi-phase CT images, making it a relatively small-scale medical dataset, we replaced the ResNet50 image encoder with a lighter ResNet18 backbone. ResNet18, with its reduced model complexity and parameter count, is more appropriate for limited-data scenarios. To ensure a fair comparison, the same training settings were applied.

As shown in Table 4, training ResNet18 from scratch (Model 1) significantly improved the average accuracy to 76.92 ± 4.41% and the AUC to 0.89 ± 0.02. Compared to using ResNet50 as the image encoder in Model 1, this represents a 5.25% increase in average accuracy and a 0.01 improvement in AUC. Notably, the classification accuracy for the HEM category improved substantially by 23.40%. Furthermore, initializing ResNet18 with ImageNet pretraining before fine-tuning (Model 2) led to a further increase in performance, with the average accuracy reaching 79.27±3.06% and the AUC improving to 0.91±0.03. Compared to the ResNet50-based Model 2, this represents a 5.19% gain in average accuracy and a 0.04 improvement in AUC.

**Table 4.** Results on the MPCT-FLLs Dataset Using ResNet-18 as the Image Encoder

|  | CYST | FNH | HCC | HEM | Avg. Acc | AUC |
| --- | --- | --- | --- | --- | --- | --- |
| Model1 | **96.25±2.15** | 72.39±18.10 | 67.35±12.21 | **67.37±20.63** | 76.92±4.41 | 0.89±0.02 |
| Model2 | 95.43±1.95 | **82.44±11.97** | **76.80±6.60** | 60.78±10.28 | **79.27±3.06** | **0.91±0.03** |

## 5 Conclusion and Future Work

In this paper, we propose the Liver-VLM model for FLLs classification in multi-phase CT imaging. By incorporating class-level text prompts into the classification process, the model enhances image-text alignment and improves classification performance under limited annotation settings. Experimental results on the MPCT-FLLs dataset demonstrate that Liver-VLM outperforms both the standard CLIP and MedCLIP models in terms of average accuracy and AUC, as shown in Table 3. Further ablation studies in Table 4 show that the lightweight ResNet18 backbone achieves better performance than the deeper ResNet50. Notably, training ResNet18 from scratch or with ImageNet pretraining both improved average accuracy and AUC, indicating its suitability for limited-data medical scenarios.

These findings underscore the effectiveness of combining multimodal pretraining with supervised learning to improve feature representations and diagnostic performance in liver lesion classification. However, despite the encouraging improvements, the average classification accuracy remains insufficient for reliable clinical application. Enhancing diagnostic precision and robustness will therefore be an important direction for future research.

## 6  Acknowledgement

Authors would like to thank Mr. Ruibo Hou of Ritsumeikan University, Japan for his helpful advice on this research and Dr. Rahul Jain of Ritsumeikan University, Japan for his kind English proof. This research was supported in part by Natural Science Foundation of Xiamen City, Fujian Province, China under the Grant No. 3502Z20227199 and in part by the Grant-in Aid for Scientific Research from the Japanese Ministry for Education, Science, Culture and Sports (MEXT) under the Grant No. 20KK0234 and No. 21H03470.

6. Song, J., Dong, H., Chen, Y., Zhang, X., Zhan, G., Jain, R.K., Chen, Y.-W.: Early Recurrence Prediction of Hepatocellular Carcinoma Using Deep Learning Frameworks with Multi-Task Pre-Training. Information 15(8), 493 (2024). doi:10.3390/info15080493.
7. Desai, K., Johnson, J.: VirTex: Learning Visual Representations from Textual Annotations. In: CVPR 2021, pp. 11162–11172. IEEE, Nashville (2021).
8. Sariyildiz, B., Nazarieh, S., Ricci, E.: ICMLM: Image Conditioned Masked Language Modeling. In: ACM MM 2020, pp. 2590–2598. ACM, Seattle (2020).
9. Zhang, Y., Jiang, H., Miura, Y., Manning, C.D., Langlotz, C.P.: Contrastive Learning of Medical Visual Representations from Paired Images and Text. In: arXiv preprint arXiv:2010.00747 (2020)
10. Radford, A., Kim, J., Hallacy, C., Ramesh, A., Goh, G., Agarwal, S., Sastry, G., Askell, A., Mishkin, P., Clark, J., Krueger, G., Sutskever, I.: Learning Transferable Visual Models From Natural Language Supervision. In: ICML 2021, pp. 8748–8763. PMLR, Virtual (2021).
11. Li, J., Selvaraju, R. R., Gotmare, A., Joty, S., Xiong, C., Hoi, S. C.: BLIP: Bootstrapped Language-Image Pre-training for Unified Vision-Language Understanding and Generation. arXiv preprint arXiv:2201.12086 (2022).
12. Liu, H., Peng, H., Yu, Z., Ma, X., Wang, J., Ma, H., Wang, Y., Wu, J., Xie, S.: Visual Instruction Tuning. arXiv preprint arXiv:2304.08485 (2023).
13. Wang, Z., Yang, J., Wang, D., Xu, Y., Bai, J., Zhou, S.K.: MedCLIP: Contrastive Learning from Unpaired Medical Images and Text. In: MICCAI 2022, LNCS, vol. 13433, pp. 40–50. Springer, Singapore (2022).
14. He, K., Zhang, X., Ren, S., Sun, J.: Deep Residual Learning for Image Recognition. In: CVPR 2016, pp. 770–778. IEEE, Las Vegas (2016).
15. Devlin, J., Chang, M.W., Lee, K., Toutanova, K.: BERT: Pre-training of Deep Bidirectional Transformers for Language Understanding. In: NAACL-HLT 2019, pp. 4171–4186. ACL, Minneapolis (2019).
16. Xu, Y., et al.: PA-ResSeg: A Phase Attention Residual Network for Liver Tumor Segmentation from Multi-phase CT Images. In: Medical Physics, vol. 48, no. 7, pp. 3752–3766. AAPM, (2021)
17. Loshchilov, I., Hutter, F.: Decoupled Weight Decay Regularization. In: ICLR 2019. OpenReview, New Orleans (2019).